\documentclass{article}

    \usepackage[nonatbib,final]{neurips_2022}
    \usepackage[numbers]{natbib}

\usepackage[utf8]{inputenc} 
\usepackage[T1]{fontenc}    
\usepackage{hyperref}       
\usepackage{url}            
\usepackage{booktabs}       
\usepackage{amsfonts}       
\usepackage{nicefrac}       
\usepackage{microtype}      
\usepackage{xcolor}         

\usepackage{amsmath,amsfonts,bm}

\def\eqref#1{equation~\ref{#1}}

\def\1{\bm{1}}

\DeclareMathAlphabet{\mathsfit}{\encodingdefault}{\sfdefault}{m}{sl}
\SetMathAlphabet{\mathsfit}{bold}{\encodingdefault}{\sfdefault}{bx}{n}

\def\gE{{\mathcal{E}}}

\def\gG{{\mathcal{G}}}

\def\gR{{\mathcal{R}}}

\def\gV{{\mathcal{V}}}

\usepackage{amsmath}
\usepackage{graphicx}
\definecolor{darkred}{HTML}{bb0000}
\definecolor{myblue}{HTML}{0096c5} 
\definecolor{RoyalBlue}{HTML}{0272bb}
\definecolor{mypurple}{HTML}{8700e0} 
\definecolor{myorange}{HTML}{d84800} 
\definecolor{ForestGreen}{RGB}{34,139,34}
\hypersetup{
  colorlinks   = true, 
  urlcolor     = RoyalBlue, 
  linkcolor    = RoyalBlue, 
  citecolor   = RoyalBlue 
}

\usepackage{graphicx}
\usepackage{xcolor}
\usepackage{multirow}
\usepackage{makecell}
\usepackage{arydshln}
\usepackage{booktabs}
\usepackage{xspace}
\usepackage{wrapfig}
\usepackage{changes}
\usepackage[nomessages]{fp}

\newcommand{\methodname}{\textsc{Dragon}\xspace}
\newcommand{\fullmethodname}{\textbf{D}eep Bidi\textbf{r}ectional L\textbf{a}n\textbf{g}uage-Kn\textbf{o}wledge Graph Pretrai\textbf{n}ing\xspace}

\newcommand\heading[1]{\textbf{#1~~}}
\newcommand\scalealign[2]{
\\[-0.3em]
\FPeval{\result}{(1/#1)}
\scalebox{#1}{\parbox{\result\linewidth}{
\begin{align}
#2
\end{align}
}}\vspace{-0.1em}
}

\title{{Deep Bidirectional Language-Knowledge Graph Pretraining}}

\author{Michihiro Yasunaga$^{1}$ ~ Antoine Bosselut$^{2}$ ~ Hongyu Ren$^{1}$ ~ Xikun Zhang$^{1}$\\
\textbf{Christopher D Manning$^{1}$ ~ Percy Liang$^{1*}$}
~ \textbf{Jure Leskovec$^{1*}$}
~\\
$^{1}$Stanford University ~ $^{2}$EPFL ~ $^{*}$Equal senior authorship\\
\scalebox{0.87}[0.9]{{\tt \{myasu,antoineb,hyren,xikunz2,manning,pliang,jure\}@cs.stanford.edu}}}

\begin{document}

\maketitle

\begin{abstract}
Pretraining a language model (LM) on text has been shown to help various downstream NLP tasks. Recent works show that a knowledge graph (KG) can complement text data, offering structured background knowledge that provides a useful scaffold for reasoning. However, these works are not pretrained to learn a deep fusion of the two modalities at scale, limiting the potential to acquire fully joint representations of text and KG. Here we propose \textbf{\methodname} (\fullmethodname), a self-supervised method to pretrain a deeply joint language-knowledge foundation model from text and KG at scale. Specifically, our model takes pairs of text segments and relevant KG subgraphs as input and bidirectionally fuses information from both modalities. We pretrain this model by unifying two self-supervised reasoning tasks, masked language modeling and KG link prediction. \methodname outperforms existing LM and LM+KG models on diverse downstream tasks including question answering across general and biomedical domains, with +5\% absolute gain on average. In particular, \methodname achieves strong performance on complex reasoning about language and knowledge (+10\% on questions involving long contexts or multi-step reasoning) and low-resource QA (+8\% on OBQA and RiddleSense), and new state-of-the-art results on various BioNLP tasks.
Our code and trained models are available at \url{https://github.com/michiyasunaga/dragon}.

\end{abstract}

\section{Introduction}
\label{sec:intro}

Pretraining learns self-supervised representations from massive raw data to help various downstream tasks \cite{bommasani2021opportunities}. Language models (LMs) pretrained on large amounts of text data, such as BERT \cite{devlin2018bert} and GPTs \cite{brown2020language}, have shown strong performance on many natural language processing (NLP) tasks. The success of these models comes from deeply interactive (contextualized) representations of input tokens learned at scale via self-supervision \cite{devlin2018bert, peters2018deep}.
Meanwhile, large knowledge graphs (KGs), such as Freebase \citep{bollacker2008freebase}, Wikidata \citep{wikidata} and ConceptNet \citep{speer2016conceptnet}, can provide complementary information to text data. KGs offer structured background knowledge by representing entities as nodes and relations between them as edges, and also offer scaffolds for structured, multi-step reasoning about entities \citep{yasunaga2021qa, zhang2022greaselm, ren2020query2box, ren2021lego} (\S \ref{sec:analysis-reasoning}).
The dual strengths of text data and KGs motivate research in pretraining deeply interactive representations of the two modalities at scale.

How to effectively combine text and KGs for pretraining is an open problem and presents challenges. Given text and KG, we need both (i) a \emph{deeply bidirectional} model for the two modalities to interact, and (ii) a \emph{self-supervised} objective to learn joint reasoning over text and KG at scale.
Several existing works \cite{zhang2019ernie, xiong2020pretrained, wang2021kepler, Agarwal2021KnowledgeGB, sun2021ernie} propose methods for self-supervised pretraining, but they fuse text and KG in a shallow or uni-directional manner.
Another line of work \cite{yasunaga2021qa, zhang2022greaselm} proposes bidirectional models for text and KG, but these models focus on finetuning on labeled downstream tasks and do not perform self-supervised learning.
Consequently, existing methods may have limited their potential to model and learn deep interactions over text and KG.

\begin{figure}
    \vspace{-7mm}
    \includegraphics[width=0.9999\textwidth]{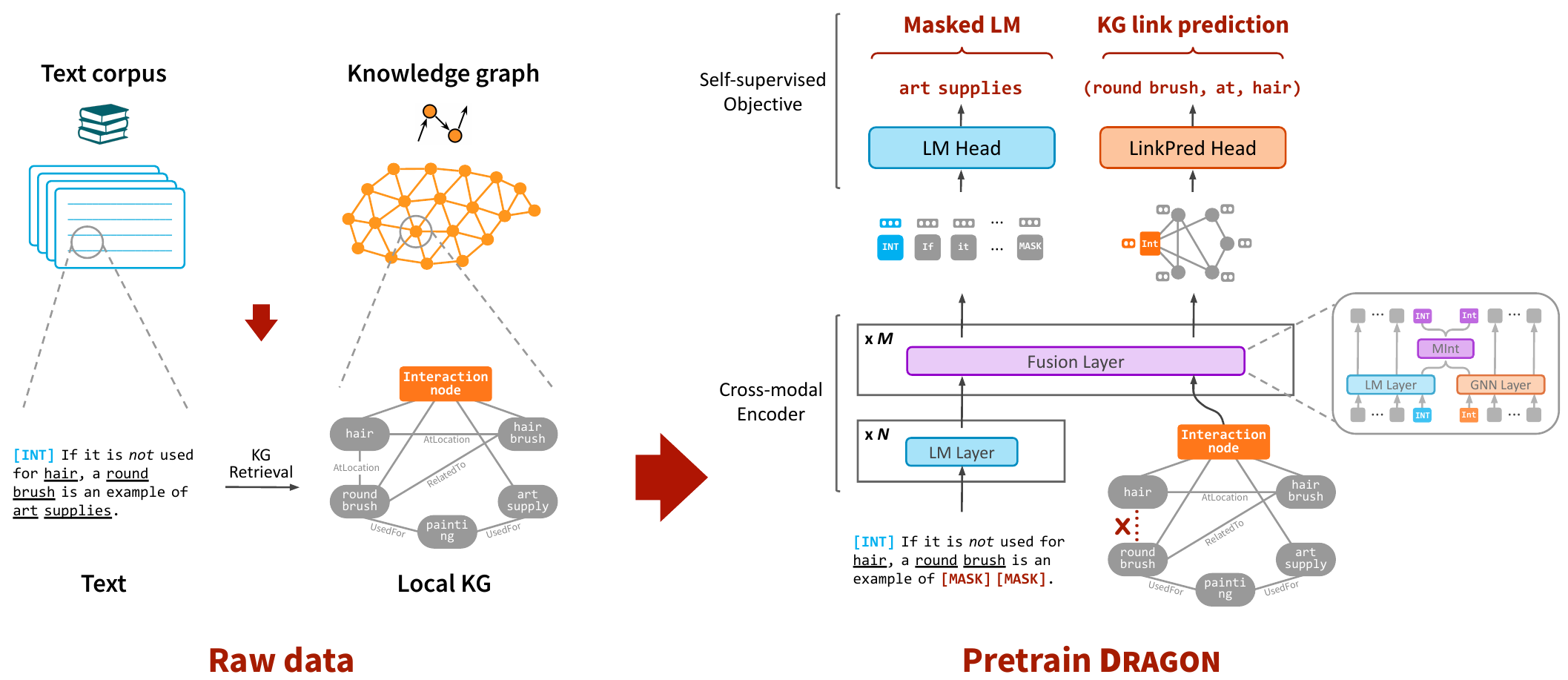}
    \vspace{-4mm}
    \caption{\footnotesize
    \textbf{Overview of our approach, \methodname}. 
    \textit{\textbf{Left}}: Given raw data of a text corpus and a large knowledge graph, we create aligned (text, local KG) pairs by sampling a text segment from the corpus and extracting a relevant subgraph from the KG (\S \ref{sec:method-input}). As the structured knowledge in KG can ground the text 
    and the text can provide the KG with rich context for reasoning,
    we aim to pretrain a language-knowledge model jointly from the text-KG pairs (\methodname).
    \textit{\textbf{Right}}: To model the interactions over text and KG, \methodname uses a cross-modal encoder that bidirectionally exchanges information between them to produce fused text token and KG node representations (\S \ref{sec:method-enc}). To pretrain \methodname jointly on text and KG, we unify two self-supervised reasoning tasks: (1) masked language modeling, which masks some tokens in the input text and then predicts them, and (2) link prediction, which holds out some edges from the input KG and then predicts them. This joint objective encourages text and KG to mutually inform each other, facilitating the model to learn joint reasoning over text and KG (\S \ref{sec:method-pretrain}).
    }
    \label{fig:overview}
    \vspace{-1mm}
\end{figure}

To address both of the above challenges and fully unify the strengths of text and KG, we propose \textbf{\methodname} (\fullmethodname), an approach that performs deeply bidirectional, self-supervised pretraining of a language-knowledge model from text and KG. 
\methodname has two core components: a cross-modal model that bidirectionally fuses text and KG, and a bidirectional self-supervised objective that learns joint reasoning over text and KG.
Concretely, as in Figure \ref{fig:overview}, we take a text corpus and a KG as raw data, and create inputs for the model by sampling a text segment from the corpus and extracting a relevant subgraph from the KG via entity linking, obtaining a (\textit{text}, \textit{local KG}) pair. 
We use a cross-modal model to encode this input into fused representations, where each layer of the model encodes the text with an LM and the KG with a graph neural network (GNN), and fuses the two with a bidirectional modality interaction module (GreaseLM \cite{zhang2022greaselm}).
We pretrain this model by unifying two self-supervised reasoning tasks: (1) masked language modeling (MLM), which masks and predicts tokens in the input text, and (2) link prediction, which drops and predicts edges in the input KG.
The intuition is that by combining the two tasks, MLM makes the model use the text jointly with structured knowledge in the KG to reason about masked tokens in the text (e.g., in Figure \ref{fig:overview}, using the ``round brush''--``art supply'' multi-hop path from the KG helps), and link prediction makes the model use the KG structure jointly with the textual context to reason about missing links in the KG (e.g., recognizing that ``round brush could be used for hair'' from the text helps).
This joint objective thus enables text to be grounded by KG structure and KG to be contextualized by text simultaneously, producing a deeply-unified language-knowledge pretrained model where information flows bidirectionally between text and KG for reasoning.

We pretrain \methodname in two domains: a general domain, using the Book corpus and ConceptNet KG \cite{speer2016conceptnet} (\S \ref{sec:experiments}), and a biomedical domain, using the PubMed corpus and UMLS KG \cite{bodenreider2004unified} (\S \ref{sec:experiments-biomed}). 
We show that \methodname improves on existing LM and LM+KG models on diverse downstream tasks across domains. For the general domain, \methodname outperforms RoBERTa \cite{liu2019roberta}, our base LM without KGs, on various commonsense reasoning tasks such as CSQA, OBQA, RiddleSense and HellaSwag, with +8\% absolute accuracy gain on average.
For the biomedical domain, \methodname improves on the previous best LM, BioLinkBERT \cite{yasunaga2022linkbert}, and sets a new state of the art on BioNLP tasks such as MedQA and PubMedQA, with +3\% accuracy gain.
In particular, \methodname exhibits notable improvements on QA tasks involving complex reasoning (+10\% gain on multi-step, negation, hedge, or long context reasoning) and on downstream tasks with limited training data (+8\% gain).
These results show that our deep bidirectional self-supervision over text and KG produces significantly improved language-knowledge representations compared to existing models.

\subsection{Related work}
\label{sec:related}

\heading{Knowledge-augmented LM pretraining.}
Knowledge integration is active research for improving LMs. One line of works is retrieval-augmented LMs \cite{guu2020realm, lewis2020retrieval, borgeaud2021improving}, which retrieve relevant text from a corpus and integrate it into LMs as additional knowledge. Orthogonal to these works, we focus on using knowledge bases as background knowledge, to ground reasoning about entities and facts.

Closest to our work are works that integrate knowledge bases in LM pretraining.
One line of research aims to add entity features to LMs \cite{zhang2019ernie, Peters2019KnowledgeEC, rosset2020knowledge};
Some works use the KG entity information or structure to create additional training signals \cite{xiong2020pretrained, Shen2020ExploitingSK, wang2021kepler, liu2020self, Yu2022JAKETJP, ke2021jointgt};
Several works add KG triplet information directly to the LM input \cite{Liu2020KBERTEL, sun2021ernie, Agarwal2021KnowledgeGB, sun2020colake, he2020integrating}.
While these methods have achieved substantial progress, they typically propagate information between text and KG in a shallow or uni-directional (e.g., KG to text) manner, which might limit the potential to perform fully joint reasoning over the two modalities. To improve on the above works, we propose to bidirectionally interact text and KG via a deep cross-modal model and joint self-supervision, so that text and KG are grounded and contextualized by each other. We find that this improves model performance on various reasoning tasks (\S \ref{sec:experiments}).
Another distinction is that existing works in this space typically focus on adding entity- or triplet-level knowledge from KGs to LMs, and focus on solving entity/relation classification tasks. Our work significantly expands this scope in that we use larger KG subgraphs (200 nodes) as input to enable richer contextualization between KG and text, and we achieve performance improvements on a broader set of NLP tasks including QA, reasoning and text classification tasks.

\heading{KG-augmented question answering.}
Various works designed KG-augmented reasoning models for question answering \cite{lin2019kagnet, feng2020scalable, lv2020graph, wang2021gnn, mihaylov2018knowledgeable, yang2019enhancing, sun2018open, sun2019pullnet, yan2021learning, sun2022jointlk, xu2021human}. In particular, recent works such as QAGNN \cite{yasunaga2021qa} and GreaseLM \cite{zhang2022greaselm} suggest that a KG can scaffold reasoning about entities with its graph structure, and help for complex question answering (e.g., negation, multi-hop reasoning).
These works typically focus on training or finetuning models on particular QA datasets. In contrast, we generalize this and integrate KG-augmented reasoning into general-purpose pretraining. Our motivation is that self-supervised pretraining allows the model to learn from larger and more diverse data, helping to learn richer interactions between text and KGs and to acquire more diverse reasoning abilities beyond specific QA tasks.
We find that our proposed pretraining approach (\methodname) offers significant boosts over the baseline QA models (e.g.~GreaseLM) on diverse downstream tasks (\S \ref{sec:experiments}). This opens a new research avenue in scaling up various carefully-designed QA models to pretraining.

\heading{KG representation learning.}
Our link prediction task used in pretraining is motivated by research in KG representation learning. Link prediction is a fundamental task in KGs \cite{trouillon2016complex, kazemi2018simple}, and various works study methods to learn KG entity and relation embeddings for link prediction, such as TransE \cite{bordes2013translating}, DistMult \cite{yang2015embedding} and RotatE \cite{sun2019rotate}.
Several works additionally use textual data or pretrained LMs to help learn KG embeddings and link prediction \cite{riedel2013relation, toutanova2015representing, xie2016representation, yao2019kg, kim2020multi, li2022multitask}.
While these works focus on the KG-side representations, we extend the scope and use the KG-side objective (link prediction) jointly with a text-side objective (language modeling) to train a mutually-interactive text-KG model.

\newcommand{\inttoken}{w_{\rm{int}}}
\newcommand{\intnode}{v_{\rm{int}}}
\newcommand{\fenc}{f_\text{enc}}
\newcommand{\fhead}{f_\text{head}}
\newcommand\mask{\texttt{[MASK]}\xspace}
\newcommand\cls{\texttt{[CLS]}\xspace}

\section{\fullmethodname (\methodname)}
\label{sec:method}
We propose \methodname, an approach that performs deeply bidirectional, self-supervised pretraining of a language-knowledge model from text and KG.
Specifically, as illustrated in Figure \ref{fig:overview}, we take a text corpus and a large knowledge graph as raw data, and create input instances for the model by sampling coarsely-aligned (text segment, local KG) pairs (\S \ref{sec:method-input}). To learn mutual interactions over text and KG, \methodname consists of a cross-modal encoder (GreaseLM) that fuses the input text-KG pair bidirectionally (\S \ref{sec:method-enc}), and a pretraining objective that performs bidirectional self-supervision on the text-KG input (\S \ref{sec:method-pretrain}). 
Our pretraining objective unifies masked language modeling (MLM) and KG link prediction (LinkPred) to make text and KG mutually inform each other and learn joint reasoning over them.
Finally, we describe how we finetune the pretrained \methodname model for downstream tasks (\S \ref{sec:method-finetune}).
While each individual piece of our approach (GreaseLM, MLM, LinkPred) is not new in itself, we are the first to bring them together effectively and demonstrate that the resulting model has strong empirical results. (\S \ref{sec:experiments}, \S \ref{sec:experiments-biomed}).

\heading{Definitions.}
We define a text corpus $\mathcal{W}$ as a set of text segments $\mathcal{W} = \{W\}$, and each text segment $W$ as a sequence of tokens (words), $W = (w_1, ..., w_I)$.
We define a knowledge graph (KG) as a multi-relational graph $\mathcal{G} = (\mathcal{V}, \mathcal{E})$, where $\gV$ is the set of entity nodes in the KG and $\gE \subseteq$ $\gV \times \gR \times \gV$ is the set of edges (triplets) that connect nodes in $\gV$, with $\gR$ being the set of relation types $\{r\}$. Each triplet $(h,r,t)$ in a KG can represent a knowledge fact such as $(\texttt{Paris}, \texttt{in}, \texttt{France})$.
As a raw KG is often large, with millions of nodes, a subgraph of the raw KG (\textit{local KG}) is considered: $G=(V, E)$ where $V = \{v_1, ..., v_J\} \subseteq \gV$ and $E\subseteq \gE$. 
We define a language-knowledge model to be a composition of two functions, $\fhead(\fenc(X))$, where the encoder $\fenc$ takes in an input $X =$ $(\text{text segment } W, \text{local KG } G)$, and produces a contextualized vector representation for each text token, $(\mathbf{H}_1, ..., \mathbf{H}_I)$, and for each KG node, $(\mathbf{V}_1, ..., \mathbf{V}_J)$. A language model is a special case of a language-knowledge model with no KG ($J=0$).
The head $\fhead$ uses these representations to perform self-supervised tasks in the pretraining step and to perform downstream tasks in the finetuning step.

\subsection{Input representation}
\label{sec:method-input}
Given a text corpus $\mathcal{W}$ and a large knowledge graph $\mathcal{G}$, we create input instances for the model by preparing (text segment $W$, local KG $G$) pairs. We want each pair's text and KG to be (roughly) semantically aligned so that the text and KG can mutually inform each other and facilitate the model to learn interactive reasoning between the two modalities.
Specifically, for each text segment $W$ from $\mathcal{W}$, we extract a relevant local KG $G$ for it from $\mathcal{G}$ via the following KG retrieval process.

\heading{KG retrieval.}
Given a text segment $W$, we link entity mentions in $W$ to entity nodes in $\mathcal{G}$ to get an initial set of nodes $V_{\text{el}}$. We then add their 2-hop bridge nodes from $\mathcal{G}$ to get the total retrieved nodes $V \subseteq \mathcal{V}$. Lastly, we add all edges that span these nodes in $\mathcal{G}$ to get $E \subseteq \mathcal{E}$, which yields the final local KG, $G=(V,E)$, as well as our final input instance $X=(W,G)$. Appendix \ref{app:setup:linking} provides more details on KG retrieval.
Henceforth, we use ``KG'' to refer to this local KG $G$ unless noted otherwise.

\heading{Modality interaction token/node.}
For each resulting (text, KG) pair, we further add a special token (interaction token) $\inttoken$ to the text and a special node (interaction node) $\intnode$ to the KG, which will serve as an information pooling point for each modality as well as an interface for modality interaction in our cross-modal encoder (\S \ref{sec:method-enc}).
Specifically, we prepend $\inttoken$ to the original text $W \!=\! (w_1, ..., w_I)$, and connect $\intnode$ to the entity-linked nodes in the original KG, $V_{\text{el}} \subseteq V \!=\! \{v_1, ..., v_J\}$, using a new relation type $r_{\text{el}}$.
The interaction token and node can also be used to produce a pooled representation of the whole input, e.g., when finetuning for classification tasks (\S \ref{sec:method-finetune}).

\subsection{Cross-modal encoder}
\label{sec:method-enc}
To model mutual interactions over the text and KG, we use a bidirectional sequence-graph encoder for $\fenc$ which takes in the text tokens and KG nodes and exchanges information across them for multiple layers to produce a fused representation of each token and node (Figure \ref{fig:overview} right):
\scalealign{0.9}{
    (\mathbf{H}_{\rm{int}}, \mathbf{H}_1, ..., \mathbf{H}_I), (\mathbf{V}_{\rm{int}}, \mathbf{V}_1, ..., \mathbf{V}_J)  = \fenc((\inttoken, w_1, ..., w_I), (\intnode, v_1, ..., v_J))
}
While we may use any deep bidirectional sequence-graph encoder for $\fenc$, for controlled comparison with existing works, we adopt the existing top-performing sequence-graph architecture, GreaseLM \cite{zhang2022greaselm}, which combines Transformers \cite{vaswani2017attention} and graph neural networks (GNNs) to fuse text-KG inputs.

Specifically, GreaseLM first uses $N$ layers of Transformer language model (LM) layers to map the input text into initial token representations, and uses KG node embeddings to map the input KG nodes into initial node representations,
\scalealign{0.85}{
    (\mathbf{H}_{\rm{int}}^{(0)}, \mathbf{H}_1^{(0)}, ..., \mathbf{H}_I^{(0)}) &= \text{LM-Layers}(\inttoken, w_1 ..., w_I),\\
    (\mathbf{V}_{\rm{int}}^{(0)}, \mathbf{V}_1^{(0)}, ..., \mathbf{V}_J^{(0)}) &= \text{Node-Embedding}(\intnode, v_1, ..., v_J).
}
Then it uses $M$ layers of text-KG fusion layers to encode these token/node representations jointly into the final token/node representations,
\scalealign{0.85}{
    (\mathbf{H}_{\rm{int}}, ..., \mathbf{H}_I), (\mathbf{V}_{\rm{int}}, ..., \mathbf{V}_J)  = \text{Fusion-Layers}((\mathbf{H}_{\rm{int}}^{(0)}, ..., \mathbf{H}_I^{(0)}), (\mathbf{V}_{\rm{int}}^{(0)}, ..., \mathbf{V}_J^{(0)})),
}
where each of the fusion layers ($\ell \!=\! 1,...,M$) performs the following:
\scalealign{0.85}{
    (\widetilde{\mathbf{H}}_{\rm{int}}^{(\ell)}, \mathbf{H}_1^{(\ell)}, ..., \mathbf{H}_I^{(\ell)}) & = \text{LM-Layer}(\mathbf{H}_{\rm{int}}^{(\ell-1)}, \mathbf{H}_1^{(\ell-1)}, ..., \mathbf{H}_I^{(\ell-1)}),\\
    (\widetilde{\mathbf{V}}_{\rm{int}}^{(\ell)}, \mathbf{V}_1^{(\ell)}, ..., \mathbf{V}_J^{(\ell)}) & = \text{GNN-Layer}(\mathbf{V}_{\rm{int}}^{(\ell-1)}, \mathbf{V}_1^{(\ell-1)}, ..., \mathbf{V}_J^{(\ell-1)}),\\
    [{\mathbf{H}}_{\rm{int}}^{(\ell)}; {\mathbf{V}}_{\rm{int}}^{(\ell)}] & = \text{MInt}([\widetilde{\mathbf{H}}_{\rm{int}}^{(\ell)}; \widetilde{\mathbf{V}}_{\rm{int}}^{(\ell)}]).
}
Here GNN induces graph structure-aware representations of KG nodes, $[\cdot\,;\cdot]$ does concatenation, and \text{MInt} (modality interaction module) exchanges information between the interaction token (text side) and interaction node (KG side) via an MLP.
For more details on GreaseLM, we refer readers to \cite{zhang2022greaselm}.

\subsection{Pretraining objective}
\label{sec:method-pretrain}
We aim to pretrain the \methodname model so that it learns joint reasoning over text and a KG.
To ensure that the text and KG mutually inform each other and the model learns bidirectional information flow, we unify two self-supervised reasoning tasks: masked language modeling and KG link prediction.

\heading{Masked language modeling (MLM).}
MLM is a common pretraining task used for language models (e.g., BERT \cite{devlin2018bert}, RoBERTa \cite{liu2019roberta}), which masks some tokens in the input text and predicts them. 
This task makes the model use non-masked context to reason about masked tokens, and in particular, as our approach takes a joint text-KG pair as input, we expect that MLM can encourage the model to learn to use the text \textit{jointly with} structured knowledge in the KG to reason about masks in the text (e.g., in the example of Figure \ref{fig:overview}, besides the textual context, recognizing the ``round brush''--``art supply'' path from the KG can help together to predict the masked tokens ``art supplies'').

Concretely, to perform the MLM task, we mask a subset of tokens in the input text, $M \subseteq W$, with a special token \mask, and let the task head $\fhead$ be a linear layer that takes the contextualized token vectors $\{\mathbf{H}_{i}\}$ from the encoder to predict the original tokens. The objective is a cross-entropy loss:
\scalealign{0.95}{
\mathcal{L}_{\text{MLM}} = -\sum_{i\in {M}} \log p(w_{i} \mid \mathbf{H}_{i}).
}

\heading{Link prediction (LinkPred).}
While the MLM task predicts for the text side, link prediction holds out some edges and predicts them for the input KG. Link prediction is a fundamental task in KGs \cite{sun2019rotate} and makes the model use the structure of KGs to perform reasoning (e.g., using a compositional path ``X's mother's husband is Y'' to deduce a missing link ``X's father is Y''). In particular, as our approach takes a joint text-KG pair as input, we expect that link prediction can encourage the model to learn to use the KG structure \textit{jointly with} the textual context to reason about missing links in the KG (e.g., in Figure \ref{fig:overview}, besides the KG structure, recognizing that ``round brush could be used for hair'' from the text can help together to predict the held-out edge \texttt{(round\_brush,\!\! at,\!\! hair)}).

Concretely, to perform the link prediction task, we hold out a subset of edge triplets from the input KG, $S = \{(h,r,t)\} \subseteq E$. 
For the task head $\fhead$, we adopt a KG representation learning framework, which maps each entity node ($h$ or $t$) and relation ($r$) in the KG to a vector, $\mathbf{h}, \mathbf{t}, \mathbf{r}$, and defines a scoring function $\phi_r(\mathbf{h}, \mathbf{t})$ to model positive/negative triplets.
Specifically, we let $\mathbf{h} = \mathbf{V}_h$, $\mathbf{t} = \mathbf{V}_t$, $\mathbf{r} = \mathbf{R}_r$, with $\{\mathbf{V}_j\}$ being the contextualized node vectors from the encoder, and $\mathbf{R} = \{\mathbf{r}_1, ..., \mathbf{r}_{|\mathcal{R}|}\}$ being learnable relation embeddings.
We consider a KG triplet scoring function $\phi_r(\mathbf{h}, \mathbf{t})$ such as
\scalealign{0.95}{
\label{eq:kg_scoring}
    \text{DistMult \cite{yang2015embedding}:~} \langle\mathbf{h}, \mathbf{r}, \mathbf{t}\rangle,~~~~
    \text{TransE \cite{bordes2013translating}:~} - \|\mathbf{h}+\mathbf{r}-\mathbf{t}\|,~~~~
    \text{RotatE \cite{sun2019rotate}:~} - \|\mathbf{h} \odot \mathbf{r}-\mathbf{t}\|,
}
where $\langle \cdot, \cdot, \cdot \rangle$ denotes the trilinear dot product and $\odot$ the Hadamard product. A higher $\phi$ indicates a higher chance of $(h,r,t)$ being a positive triplet (edge) instead of negative (no edge). We analyze the choices of scoring functions in \S \ref{sec:exp-ablation}.
For training, we optimize the objective:
\scalealign{0.95}{
    \label{eq:linkpred-loss}
    \mathcal{L}_{\text{LinkPred}}=\sum_{(h,r,t)\in {S}} \left( -\log{\sigma(\phi_r(\mathbf{h}, \mathbf{t}) +\gamma)}
    +\frac{1}{n}\sum_{(h', r, t')}\log{\sigma(\phi_r(\mathbf{h'}, \mathbf{t'}) +\gamma)} \right),
}
where $(h', r, t')$ are $n$ negative samples corresponding to the positive triplet $(h, r, t)$, $\gamma$ is the margin, and $\sigma$ is the sigmoid function.
The intuition of this objective is to make the model predict triplets of the held-out edges $S$ as positive and other random triplets as negative.

\heading{Joint training.}
To pretrain \methodname, we optimize the MLM and LinkPred objectives jointly: $\mathcal{L} =\mathcal{L}_{\text{MLM}} +\mathcal{L}_{\text{LinkPred}}$.
This joint objective unifies the effects of MLM and LinkPred, which encourage the model to simultaneously ground text with KG structure and contextualize KG with text, facilitating bidirectional information flow between text and KGs for reasoning. We show in \S \ref{sec:exp-ablation} that the joint objective yields a more performant model than using one of the objectives alone.

\subsection{Finetuning}
\label{sec:method-finetune}
Lastly, we describe how we finetune \methodname for downstream tasks such as text classification and multiple-choice QA (MCQA). 
Given an input text $W$ (e.g., concatenation of a question and an answer choice in the case of MCQA), we follow the same steps as \S \ref{sec:method-input} and \S \ref{sec:method-enc} to retrieve a relevant local KG $G$ and encode them jointly into contextualized token/node vectors, $(\mathbf{H}_{\rm{int}}, \mathbf{H}_1, ...,$ $\mathbf{H}_I)$, $(\mathbf{V}_{\rm{int}}, \mathbf{V}_1, ..., \mathbf{V}_J)$. We then compute a pooled representation of the whole input as $\mathbf{X} = \text{MLP}(\mathbf{H}_{\rm{int}}, \mathbf{V}_{\rm{int}}, \mathbf{G})$, where $\mathbf{G}$ denotes attention-based pooling of $\{\mathbf{V}_j \mid v_j \in \{v_1, ..., v_J\}\}$ using $\mathbf{H}_{\rm{int}}$ as a query.
Finally, the pooled representation $\mathbf{X}$ is used to perform the downstream task, in the same way as how the \cls representation is used in LMs such as BERT and RoBERTa.

The difference from GreaseLM is that while GreaseLM only performs finetuning as described in this section (hence, it is an LM \textit{finetuned} with KGs), \methodname performs self-supervised pretraining as described in \S \ref{sec:method-pretrain} (hence, it can be viewed as an LM \textit{pretrained + finetuned} with KGs).

\section{Experiments: General domain}
\label{sec:experiments}
We experiment with the proposed approach \methodname in a general domain first. We pretrain \methodname using the Book corpus and ConceptNet KG (\S \ref{sec:exp-pretrain-setup}), and evaluate on diverse downstream tasks (\S \ref{sec:exp-eval}). We show that \methodname significantly improves on existing models (\S \ref{sec:exp-result}). We extensively analyze the effect of \methodname's key design choices such as self-supervision and use of KGs (\S \ref{sec:analysis-reasoning}, \ref{sec:analysis-pretrain}, \ref{sec:exp-ablation}).
We also experiment in the biomedical domain in \S \ref{sec:experiments-biomed}.

\subsection{Pretraining setup}
\label{sec:exp-pretrain-setup}

\heading{Data.}
For the text data, we use BookCorpus \cite{Zhu_2015_ICCV}, a general-domain corpus widely used in LM pretraining (e.g., BERT, RoBERTa). It has 6GB of text from online books.
For the KG data, we use {ConceptNet} \citep{speer2016conceptnet}, a general-domain knowledge graph designed to capture background commonsense knowledge. It has 800K nodes and 2M edges in total.
To create a training instance, we sample a text segment of length up to 512 tokens from the text corpus, then retrieve a relevant KG subgraph of size up to 200 nodes (details in Appendix \ref{app:setup:linking}), by which we obtain an aligned (text, local KG) pair.

\heading{Implementation.}
For our encoder (\S \ref{sec:method-enc}), we use the exact same architecture as GreaseLM \cite{zhang2022greaselm} (19 LM layers followed by 5 text-KG fusion layers; 360M parameters in total). As done by \cite{zhang2022greaselm}, we initialize parameters in the LM component with the RoBERTa-Large release \cite{liu2019roberta} and initialize the KG node embeddings with pre-computed ConceptNet entity embeddings (details in Appendix \ref{app:setup:graph}). 
For the link prediction objective (\S \ref{sec:method-pretrain}, Equation \ref{eq:linkpred-loss}), we use DistMult \cite{yang2015embedding} for KG triplet scoring, with a negative exampling of 128 triplets and a margin of $\gamma=0$.
To pretrain the model, we perform MLM with a token masking rate of 15\% and link prediction with an edge drop rate of 15\%.
We pretrain for 20,000 steps with a batch size of 8,192 and a learning rate of 2e-5 for parameters in the LM component and 3e-4 for the others. 
Training took 7 days on eight A100 GPUs using FP16.
Additional details on the hyperparameters can be found in Appendix \ref{app:setup:hyperparameters}.

\subsection{Downstream evaluation tasks}
\label{sec:exp-eval}

We finetune and evaluate \methodname on nine diverse commonsense reasoning benchmarks: 
{CommonsenseQA (\textbf{CSQA})} \cite{talmor2018commonsenseqa},
{OpenbookQA (\textbf{OBQA})} \cite{obqa}, 
{RiddleSense (\textbf{Riddle})} \cite{lin2021riddlesense},
{AI2 Reasoning Challenge\! --\! Challenge Set (\textbf{ARC})} \cite{clark2018think},
{\textbf{CosmosQA}} \cite{huang2019cosmos},
{\textbf{HellaSwag}} \cite{zellers2019hellaswag},
{Physical Interaction QA (\textbf{PIQA})} \cite{bisk2020piqa},
{Social Interaction QA (\textbf{SIQA})} \cite{sap2019socialiqa}, and
{Abductive Natural Language Inference (\textbf{aNLI})} \cite{bhagavatula2019abductive}. 
For CSQA, we follow the in-house data splits used by prior works \cite{lin2019kagnet}. For OBQA, we follow the original setting where the models only use the question as input and do not use the extra science facts.
Appendix \ref{app:setup:eval} provides the full details on these tasks and data splits.
Hyperparameters used for finetuning can be found in Appendix \ref{app:setup:hyperparameters}.

\subsection{Baselines}
\label{sec:exp-baseline}

\heading{LM.}
To study the effect of using KGs, we compare \methodname with the vanilla language model, RoBERTa \cite{liu2019roberta}. 
As we initialize \methodname's parameters using the RoBERTa-Large release (\S \ref{sec:exp-pretrain-setup}), for fair comparison, we let the baseline be such that we take the RoBERTa-Large release and continue pretraining it with the vanilla MLM objective on the same text data for the same number of steps as \methodname. Hence, the only difference is that \methodname uses KGs during pretraining while RoBERTa does not.
We then perform standard LM finetuning of RoBERTa on downstream tasks.

\heading{LM finetuned with KG.}
We also compare with existing KG-augmented QA models, QAGNN \cite{yasunaga2021qa} and GreaseLM \cite{zhang2022greaselm}, which \textit{finetune} a vanilla LM (i.e.~RoBERTa-Large) with a KG on downstream tasks, but do not \textit{pretrain} with a KG. GreaseLM is the existing top-performing model in this paradigm.
As we use the same encoder architecture as GreaseLM for \methodname, the only difference from GreaseLM is that \methodname performs self-supervised pretraining while GreaseLM does not.

\subsection{Results}
\label{sec:exp-result}

Table \ref{tab:commonsense_main} shows performance on the 9 downstream commonsense reasoning tasks. Across all tasks, \methodname consistently outperforms the existing LM (RoBERTa) and KG-augmented QA models (QAGNN, GreaseLM), e.g., +7\% absolute accuracy boost over RoBERTa and +5\% over GreaseLM on \textit{OBQA}. 
These accuracy boosts indicate the advantage of \methodname over RoBERTa (KG reasoning) and over GreaseLM (pretraining).
The gain is especially significant on datasets that have small training data such as \textit{ARC}, \textit{Riddle} and \textit{OBQA}, and datasets that require complex reasoning such as \textit{CosmosQA} and \textit{HellaSwag}, which we analyze in more detail in the following sections.

\subsubsection{Analysis: Effect of knowledge graph}
\label{sec:analysis-reasoning}
The first key contribution of \methodname (w.r.t. existing LM pretraining methods) is that we incorporate KGs.
We find that this significantly improves the model's performance for robust and complex reasoning, such as resolving multi-step reasoning and negation, as we discuss below.

\begin{table*}
\vspace{-6mm}
\centering
  \scalebox{0.8}{
  \small
  \begin{tabular}{l ccccccccc}
    \toprule
     & \textbf{CSQA} & \textbf{OBQA} & \textbf{Riddle} & \textbf{ARC} & \!\!\textbf{CosmosQA}\!\!  & \!\!\textbf{HellaSwag}\!\! & \textbf{PIQA} & \textbf{SIQA} & \textbf{aNLI}  \\
    \midrule
    RoBERTa \cite{liu2019roberta} & 68.7 & 64.9 & 60.7 & 43.0 & 80.5 & 82.3 & 79.4 & 75.9 & 82.7 
    \\
    \midrule
    QAGNN \cite{yasunaga2021qa}   & 73.4 & 67.8 & 67.0 & 44.4 & 80.7 & 82.6 & 79.6 & 75.7 & 83.0 
    \\
    GreaseLM \cite{zhang2022greaselm} & 74.2 & 66.9 & 67.2 & 44.7 & 80.6 & 82.8 & 79.6 & 75.5 & 83.3 
    \\
    \midrule
    \methodname (\textbf{Ours})
    & \textbf{76.0} & \textbf{72.0} & \textbf{71.3} & \textbf{48.6} & \textbf{82.3} & \textbf{85.2} & \textbf{81.1} & \textbf{76.8} & \textbf{84.0}  
    \\
    \bottomrule
  \end{tabular}}
  \vspace{-1mm}
  \caption{\footnotesize
  Accuracy on downstream commonsense reasoning tasks. \methodname consistently outperforms the existing LM (RoBERTa) and KG-augmented QA models (QAGNN, GreaseLM) on all tasks. The gain is especially significant on tasks that have small training data (\textit{OBQA}, \textit{Riddle}, \textit{ARC}) and tasks that require complex reasoning (\textit{CosmosQA}, \textit{HellaSwag}).
  }
  \label{tab:commonsense_main}
  \vspace{-1mm}
\end{table*}

\begin{table}[t]
    \centering
    \scalebox{0.7}{
    \small
    \begin{tabular}{l rrrrrrrr }
        \toprule
        \multirow{2}{*}{\textbf{}} & \textbf{Negation} & \textbf{Conjunction} & \textbf{Hedge}  & \multicolumn{4}{c}{\# \textbf{Prepositional Phrases}}  & \multicolumn{1}{c}{\# \textbf{Entities}}\\ 
        & \textbf{} & \textbf{} & \textbf{} & 0 & 1 & 2 & 3  & >10 \\
        \midrule
        RoBERTa   & 61.7 & 70.9 & 68.6 &  67.6 & 71.0 & 71.1 & 73.1 & 74.5 
        \\
        \midrule
        QAGNN     & 65.1 & 74.5 & 74.2 & 72.1 &	71.6 & 75.6 & 71.3 & 78.6 
        \\
        GreaseLM  & 65.1 & 74.9 & 76.6  & 75.6 & 73.8 & 74.7 & 73.6 & 79.4 
        \\
        \midrule
        \methodname (\textbf{Ours}) & \textbf{75.2} & \textbf{79.6} & \textbf{77.5}  & \textbf{79.1} &	\textbf{78.2} & \textbf{77.8} &\textbf{ 80.9} & \textbf{83.5} 
        \\
        \bottomrule
    \end{tabular}
    }
    \vspace{1mm}
    \caption{\footnotesize
    Accuracy of \methodname on \textit{CSQA} + \textit{OBQA} dev sets for \textbf{questions involving complex reasoning} such as negation terms, conjunction terms, hedge terms, prepositional phrases, and more entity mentions. 
    \methodname consistently outperforms the existing LM (RoBERTa) and KG-augmented QA models (QAGNN, GreaseLM) in these complex reasoning settings.
    }
    \label{tab:analysis}
\vspace{-4mm}
\end{table}
\heading{Quantitative analysis.}
In Table \ref{tab:analysis}, we study downstream task performance of \methodname on questions involving complex reasoning. Building on \cite{yasunaga2021qa, zhang2022greaselm}, we consider several proxies to categorize complex questions: (i) presence of negation (e.g.~\textit{no}, \textit{never}), (ii) presence of conjunction (e.g.~\textit{and}, \textit{but}), (iii) presence of hedge (e.g.~\textit{sometimes}, \textit{maybe}), (iv) number of prepositional phrases, and (v) number of entity mentions. Having negation or conjunction indicates logical multi-step reasoning, having more prepositional phrases or entity mentions indicates involving more reasoning steps or constraints, and having hedge terms indicates involving complex textual nuance.
\methodname significantly outperforms the baseline LM (RoBERTa) across all these categories (e.g., +14\% accuracy for negation), which confirms that our joint language-knowledge pretraining boosts reasoning performance.
\methodname also consistently outperforms the existing KG-augmented QA models (QAGNN, GreaseLM). We find that QAGNN and GreaseLM only improve moderately on RoBERTa for some categories like conjunction or many prepositional phrases (=2,\,3), but \methodname provides substantial boosts. This suggests that through self-supervised pretraining with larger and diverse data, \methodname has learned more general-purpose reasoning abilities than the finetuning-only models like GreaseLM.

\begin{figure}[t]
    \vspace{-2mm}
    \hspace{-3mm}
    \includegraphics[width=1.03\textwidth]{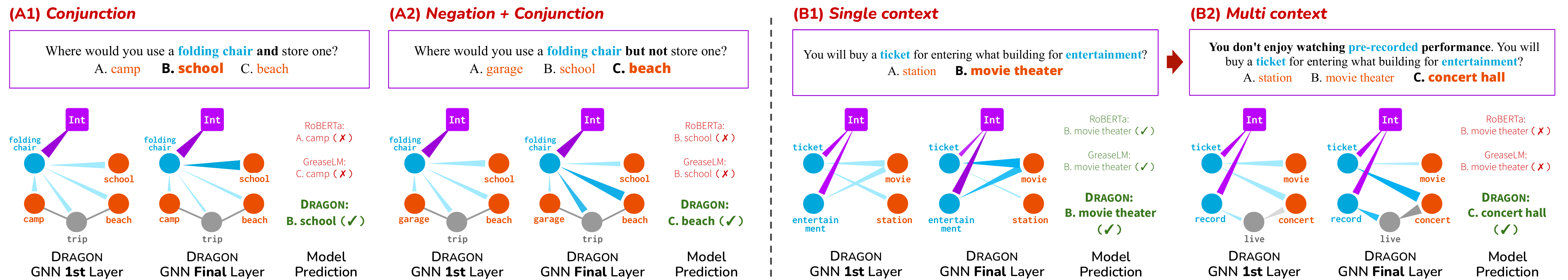}\vspace{-1mm}
    \caption{\footnotesize
    Analysis of \methodname's graph reasoning, where we visualize how graph attention weights and final predictions change given question variations. Darker and thicker edges indicate higher attention weights. \textbf{\methodname exhibits abilities to extrapolate and perform robust reasoning}. \methodname adjusts the entity attention weights and final predictions accordingly when conjunction or negation is given about entities (A1, A2) or when extra context is added to an original question (B1$\rightarrow$B2), but existing models, RoBERTa and GreaseLM, struggle to predict the correct answers.
    \textbf{A1:} \methodname's final GNN layer shows strong attention to ``school'' but weak attention to ``trip'', likely because the question states ``\textit{and} store one''---hence, the chair is \textit{not} used for a trip. 
    \textbf{A2:} \methodname shows strong attention to ``trip'' and ``beach'', likely because the question now states ``\textit{but not} store one''---hence, the chair \textit{is} used for a trip.
    \textbf{B1$\rightarrow$B2:} \methodname's final GNN layer shows strong attention to ``movie'' in the original question (B1), but after adding the extra context ``don't enjoy pre-record'' (B2), \methodname shows strong attention to ``live'' and ``concert'', leading to making the correctly adjusted prediction ``concert hall''.
    One interpretation of these findings is that \methodname leverages the KG's graph structure as a scaffold for performing complex reasoning. 
    This insight is related to recent works that provide LMs with scratch space for intermediate reasoning \cite{yasunaga2021qa, nye2021show, wei2022chain}.
    }
    \label{fig:qualitative}
    \vspace{1mm}
\end{figure}

\heading{Qualitative analysis.}
Using the \textit{CSQA} dataset, we further conducted case studies on the behavior of \methodname's KG reasoning component, where we visualize how graph attention weights change given different question variations (Figure \ref{fig:qualitative}). We find that \methodname exhibits abilities to extrapolate and perform robust reasoning. For instance, \methodname adjusts the entity attention weights and final predictions accordingly when we add conjunction or negation about entities (A1, A2) or when we add extra context to an original question (B1$\rightarrow$B2), but existing models, RoBERTa and GreaseLM, struggle to predict the correct answers. 
As these questions are more complex than ones typically seen in the \textit{CSQA} training set, our insight is that while vanilla LMs (RoBERTa) and finetuning (GreaseLM) have limitation in learning complex reasoning, KG-augmented pretraining (\methodname) helps acquire generalizable reasoning abilities that extrapolate to harder test examples.

\subsubsection{Analysis: Effect of pretraining}
\label{sec:analysis-pretrain}
\begin{table}[t]
\begin{minipage}{0.458\textwidth}
    \centering
    \scalebox{0.7}{
    \small
    \begin{tabular}{lcc}
    \toprule
    \textbf{Method} & \textbf{CosmosQA} \scalebox{0.8}{\textbf{(10\% train)}} & \textbf{PIQA} \scalebox{0.8}{\textbf{(10\% train)}}   \\
    \midrule
    {RoBERTa}  & 72.2 & 66.4 \\
    {GreaseLM} & 73.0 & 67.0 \\
    \midrule
    \methodname (\textbf{Ours}) &  \textbf{77.9} &  \textbf{72.3} \\
    \bottomrule
    \end{tabular}
    }\vspace{1.5mm}
    \caption{\footnotesize
     Performance in low-resource setting where 10\% of finetuning data is used. \methodname attains large gains, suggesting its benefit for downstream data efficiency.
    }\vspace{1mm}
    \label{tab:low-data}
    
    \centering
    \scalebox{0.7}{
    \small
    \begin{tabular}{lcc}
    \toprule
    \textbf{Method} & \textbf{CSQA} & \textbf{OBQA}   \\
    \midrule
    {GreaseLM}  & 74.2 & 66.9 \\
    {GreaseLM-Ex} & 73.9 & 66.2 \\
    \midrule
    \methodname (\textbf{Ours})  & 76.0 & 72.0  \\
    \methodname-Ex (\textbf{Ours}) & \textbf{76.3} & \textbf{72.8} \\
    \bottomrule
    \end{tabular}
    }\vspace{1.5mm}
    \caption{\footnotesize
    Downstream performance when model capacity---number of text-KG fusion layers---is increased (``-Ex''). Increased capacity does not help for the finetuning-only model (GreaseLM), but helps when pretrained (\methodname), suggesting the promise of \methodname to be further scaled up.
    }
    \label{tab:model-cap}
\end{minipage}\hfill
\begin{minipage}{0.52\textwidth}
    \centering
    \scalebox{0.65}{
    \small
    \begin{tabular}{rlcc}
     \toprule
    \textbf{Ablation Type}\!\! & \textbf{Ablation} & \!\!\textbf{CSQA}\!\! & \!\!\textbf{OBQA}\!\! \\
    \midrule
    \!\!\multirow{3}{*}{Pretraining objective}\!\! & MLM\,+\,LinkPred (\textbf{final})\!\! &  \textbf{76.0} & \textbf{72.0} \\
    & MLM only & 74.3 & 67.2 \\
    & LinkPred only & 73.8 & 66.4 \\
    \midrule
    \!\multirow{3}{*}{LinkPred head}\!\! & DistMult (\textbf{final}) & \textbf{76.0} & \textbf{72.0} \\
    &  TransE & 75.7 & 71.4 \\
    &  RotatE & 75.8 & 71.7  \\
    \midrule
    \!\multirow{2}{*}{Cross-modal model}\!\! & Bidirectional interaction (\textbf{final}) & \textbf{76.0} & \textbf{72.0} \\
     & Concatenate at end & 74.5 & 68.0 \\
    \midrule
    \!\multirow{2}{*}{KG structure}\!\! & Use graph (\textbf{final}) & \textbf{76.0} & \textbf{72.0} \\
    &  Convert to sentence & 74.7 & 70.1 \\
    \bottomrule
    \end{tabular}}
    \vspace{2mm}
    \caption{\footnotesize
    {Ablation study} of \methodname. Using joint pretraining objective MLM + LinkPred (\S \ref{sec:method-pretrain}) outperforms using one of them only. All variants of LinkPred scoring models (DistMult, TransE, RotatE) outperform the baseline without LinkPred (``MLM only''), suggesting that \methodname can be combined with various KG representation learning models. Cross-modal model with bidirectional modality interaction (\S \ref{sec:method-enc}) outperforms combining text and KG representations only at the end.
    Finally, using KG as graph outperforms converting KG as sentences, suggesting the benefit of graph structure for reasoning. 
    }
    \label{tab:ablation}
\end{minipage}\vspace{-2mm}
\end{table}

Another key contribution of \methodname (w.r.t. existing QA models like GreaseLM) is pretraining. Here we discuss when and why our pretraining is useful.
Considering the three core factors in machine learning (data, task complexity, and model capacity), pretraining helps when the available downstream task data is smaller compared to the downstream task complexity or model capacity.
Concretely, we find that \methodname is especially helpful for the following three scenarios.

\heading{Downstream tasks with limited data.}
In Table \ref{tab:commonsense_main}, we find that \methodname provides significant boosts over GreaseLM on downstream tasks with limited finetuning data available, such as \textit{ARC} (3K training instances; +4\% accuracy gain), \textit{Riddle} (3K instances; +4\% accuracy) and \textit{OBQA} (5K instances; +5\% accuracy).
For other tasks, we also experimented with a low-resource setting where 10\% of finetuning data is used (Table \ref{tab:low-data}). Here we also see that \methodname attains significant gains over GreaseLM (+5\% accuracy on \textit{PIQA}), suggesting the improved data-efficiency of \methodname.

\heading{Complex downstream tasks.}
In Table \ref{tab:commonsense_main}, we find that \methodname provides substantial gains over GreaseLM on downstream tasks involving more complex reasoning, such as \textit{CosmosQA} and \textit{HellaSwag}, where the inputs have longer context and more entities (thus bigger local KGs). For these tasks, improvements of GreaesLM over RoBERTa were small (+0.1\% on \textit{CosmosQA}), but \methodname provides substantial boosts (+1.8\%). Our insight is that through self-supervised pretraining with larger and more diverse data, \methodname has learned richer text-KG interactions than GreaseLM, enabling solving more complex downstream tasks.
Similarly, as seen in \S \ref{sec:analysis-reasoning}, \methodname also attains large gains over GreaseLM on complex questions containing negation, conjunction and prepositional phrases (Table \ref{tab:analysis}), and extrapolates to questions more complex than seen in training sets (Figure \ref{fig:qualitative}).

\heading{Increased model capacity.}
In Table \ref{tab:model-cap}, we study downstream performance when the model capacity is increased---the number of text-KG fusion layers is increased from 5 to 7---for both GreaseLM and \methodname. We find that increased capacity does not help for the finetuning-only model (GreaseLM) as was also reported in the original GreaseLM paper, but it helps when pretrained (\methodname). This result reveals that increased model capacity can actually be beneficial when combined with pretraining, and suggests the promise of \methodname to be further scaled up.

\subsubsection{Analysis: Design choices of \methodname}
\label{sec:exp-ablation}

\heading{Pretraining objective {\rm (Table \ref{tab:ablation} top).}} The first important design choice of \methodname is the joint pretraining objective: MLM + LinkPred (\S \ref{sec:method-pretrain}). Using the joint objective outperforms using MLM or LinkPred alone (+5\% accuracy on \textit{OBQA}). This suggests that having the bidirectional self-supervised tasks on text and KG facilitates the model to fuse the two modalities for reasoning.
\\[1.5mm]
\heading{Link prediction head choice {\rm (Table \ref{tab:ablation} middle 1).}} 
KG representation learning is an active area of research, and various KG triplet scoring models are proposed (Equation \ref{eq:kg_scoring}).
We hence experimented with using different scoring models for \methodname's link prediction head (\S \ref{sec:method-pretrain}). We find that while DistMult has a slight edge, all variants we tried (DistMult, TransE, RotatE) are effective, outperforming the baseline without LinkPred (``MLM only''). This result suggests the generality of \methodname and its promise to be combined with various KG representation learning techniques. 
\\[1.5mm]
\heading{Cross-modal model {\rm (Table \ref{tab:ablation} middle 2).}} 
Another core component of \methodname is the cross-modal encoder with bidirectional text-KG fusion layers (\S \ref{sec:method-enc}). We find that if we ablate them and simply concatenate text and KG representations at the end, the performance drops substantially. This result suggests that deep bidirectional fusion is crucial to model interactions over text and KG for reasoning.
\\[1.5mm]
\heading{KG structure {\rm (Table \ref{tab:ablation} bottom).}}
The final key design of \methodname is that we leverage the graph structure of KGs via a sequence-graph encoder and link prediction objective. Here we experimented with an alternative pretraining method that drops the graph structure: we convert triplets in the local KG into sentences using a template \cite{feng2020scalable}, append them to the main text input, and perform vanilla MLM pretraining.
We find that \methodname substantially outperforms this variant (+2\% accuracy on \textit{OBQA}), which suggests that the graph structure of KGs helps the model perform reasoning.
\section{Experiments: Biomedical domain}
\label{sec:experiments-biomed}
Biomedicine is a domain with extensive background knowledge \cite{brown1999medical, lipscomb2000medical, zitnik2018modeling, bommasani2021opportunities}, and experts curate various knowledge bases for it \cite{ashburner2000gene, bodenreider2004unified, wishart2018drugbank, ruiz2021identification}.
We hypothesize that these biomedical KGs can enable deeper understanding and reasoning about biomedical text.
With this motivation, we pretrain \methodname on a biomedical corpus and KG, and evaluate on biomedical downstream tasks.

\heading{Pretraining setup.}
For the text data, we use PubMed \cite{pubmed},
a widely-used corpus in biomedial LM training (e.g., BioBERT \cite{lee2020biobert}, PubmedBERT \cite{gu2020domain}). It contains the abstracts of biomedical papers on PubMed and has 21GB of text.
For the KG data, we use the Unified Medical Language System (UMLS) \cite{bodenreider2004unified}, a widely-used knowledge graph in biomedicine. It has 300K nodes and 1M edges in total.
%
For training, we follow the same procedure as the experiment in the general domain (\S \ref{sec:exp-pretrain-setup}), except that we initialize \methodname's LM component with BioLinkBERT-Large \cite{yasunaga2022linkbert}, the state-of-the-art biomedical LM, instead of RoBERTa-Large.
Note that while ``BioLinkBERT'' has ``Link'' in its name, it is not about KG links but about citation links that the model was originally pretrained with.

\setlength{\columnsep}{5mm}
\begin{wraptable}{r}{0.42\textwidth}
\vspace{-4mm}
\hspace{-1mm}
\scalebox{0.8}{
\small
\begin{tabular}{lccc}
\toprule
\textbf{Method} & \textbf{MedQA} & \!\!\!\!\textbf{PubMedQA}\!\!\!\! & \textbf{BioASQ}  \\
\midrule
{BioBERT} \cite{lee2020biobert} & 36.7 & 60.2 & 84.1\\
{PubmedBERT} \cite{gu2020domain} & 38.1 & 55.8 & 87.5\\
\midrule
{BioLinkBERT} \cite{yasunaga2022linkbert} &  {44.6} & 72.2 & 94.8 \\
{~~+ QAGNN} &  {45.0} & 72.1 & 95.0 \\
{~~+ GreaseLM} &  {45.1} & 72.4 & 94.9 \\
\methodname (\textbf{Ours}) & \textbf{47.5} & \textbf{73.4} & \textbf{96.4}\\
\bottomrule
\end{tabular}
}\vspace{-1mm}
\caption{\footnotesize
{Accuracy on biomedical NLP tasks}. \methodname outperforms all previous biomedical LMs.
}
\label{tab:bio_main}
\vspace{-10mm}
\end{wraptable}

\heading{Downstream evaluation tasks.}
We finetune and evaluate \methodname on three popular biomedical NLP and reasoning benchmarks: 
MedQA-USMLE (\textbf{MedQA}) \cite{jin2021disease}, 
\textbf{PubMedQA} \cite{jin2019pubmedqa}, and
\textbf{BioASQ} \cite{nentidis2019results}.
Appendix \ref{app:setup:eval} provides details on these tasks and data splits.

\heading{Baselines.}
We compare \methodname with the vanilla LM (BioLinkBERT) and LMs finetuned with the KG (QAGNN and GreaseLM seeded with BioLinkBERT).

\heading{Results.}
Table \ref{tab:bio_main} summarizes model performance on the downstream tasks.
Across tasks, \methodname outperforms all the existing biomedical LMs and KG-augmented QA models, e.g., +3\% absolute accuracy boost over BioLinkBERT and +2\% over GreaseLM on \textit{MedQA}, achieving new state-of-the-art performance on these tasks.
This result suggests significant efficacy of KG-augmented pretraining for improving biomedical reasoning tasks.
Combined with the results in the general commonsense domain (\S \ref{sec:exp-result}), our experiments also suggest the domain-generality of \methodname, serving as an effective pretraining method across domains with different combinations of text, KGs and seed LMs. 

\section{Conclusion}
\label{sec:conclusion}

We presented \methodname, a self-supervised pretraining method to learn a deeply bidirectional language-knowledge model from text and knowledge graphs (KGs) at scale. In both general and biomedical domains, \methodname outperforms existing language models and KG-augmented models on various NLP tasks, and exhibits strong performance on complex reasoning such as answering questions involving long context or multi-step reasoning.

One limitation of \methodname is that it is currently an encoder model (analogous to BERT) and does not perform language generation. An important future research would be to extend \methodname to generation, and advance KG-enhanced language generation \cite{ke2021jointgt, yu2022survey}.

\section*{Reproducibility}
\renewcommand\ttdefault{cmtt}
Pretrained models, code and data are available at
\url{https://github.com/michiyasunaga/dragon}.\\
Experiments are available at\\
\url{https://worksheets.codalab.org/worksheets/0xcf9cddffff864fb382e1a2f1393c8934}.

\section*{Acknowledgment}
We thank Rok Sosic, Hamed Nilforoshan, Michael Moor, Qian Huang, members of the Stanford SNAP, P-Lambda, and NLP groups, as well as our anonymous reviewers for valuable feedback.
We also gratefully acknowledge the support of HAI Google Cloud Credits 1051203844499; DARPA under Nos.
HR00112190039 (TAMI), N660011924033 (MCS); ARO under Nos. W911NF-16-1-0342 (MURI),
W911NF-16-1-0171 (DURIP); NSF under Nos. OAC-1835598 (CINES), OAC-1934578 (HDR),
CCF-1918940 (Expeditions), IIS-2030477 (RAPID), NIH under No. R56LM013365; Stanford Data
Science Initiative, Wu Tsai Neurosciences Institute, Chan Zuckerberg Biohub, Amazon, JPMorgan
Chase, Docomo, Hitachi, Intel, JD.com, KDDI, Toshiba, NEC, and UnitedHealth Group. 
The content is solely the responsibility of the authors and
does not necessarily represent the official views of the funding entities.

\bibliographystyle{unsrtnat}
\bibliography{main}

\newpage
\newpage
\appendix

\section{Ethics, limitations and risks}
\label{app:ethics}

We outline potential ethical issues with our work below. First, \methodname is a method to fuse language representations and knowledge graph representations for joint reasoning. Consequently, \methodname could reflect the same biases and toxic behaviors exhibited by language models and knowledge graphs that are used to initialize it. For example, language models have been shown to encode biases about race, gender, and other demographic attributes \citep{sheng2020towards, weidinger2021ethical} and generate toxic outputs \cite{gehman2020realtoxicityprompts}. Because \methodname is seeded with pretrained language models that often learn these patterns, it is possible to reflect them in open-world settings. Second, the ConceptNet knowledge graph \citep{speer2016conceptnet} used in this work has been shown to encode stereotypes \citep{Mehrabi2021LawyersAD}, rather than completely clean commonsense knowledge. If \methodname were used outside these standard benchmarks in conjunction with ConceptNet as a KG, it might rely on unethical relationships in its knowledge resource to arrive at conclusions. Consequently, while \methodname could be used for applications outside these standard benchmarks, we would encourage implementers to use the same precautions they would apply to other language models and methods that use noisy knowledge sources.

Another source of ethical concern is the use of the MedQA-USMLE evaluation. While we find this clinical reasoning task to be an interesting testbed for \methodname and for joint language and knowledge reasoning in general, we do not encourage users to use these models for real world clinical prediction.

Reference: \cite{zhang2022greaselm}.

\section{Experimental Setup Details}
\label{app:setup}

\subsection{KG retrieval}
\label{app:setup:linking}

Given each input text segment $W$, we follow the procedure from \citet{yasunaga2021qa} to retrieve a relevant local KG $G$ from the raw KG $\gG =(\gV,\gE)$. First, we use the entity linker from the spaCy library\footnote{\url{https://spacy.io/}} to link entity mentions in $W$ to entity nodes in $\mathcal{G}$, obtaining an initial set of nodes $V_{\text{el}}$. Second, we add any bridge entities in $\mathcal{G}$ that are in a 2-hop path between any pair of linked entities in $V_{\text{el}}$ to get the total retrieved nodes $V \subseteq \mathcal{V}$. If the number of nodes in $V$ exceeds 200, we prune $V$ by randomly sampling 200 nodes from it to be the final retrieved nodes $V$.
Lastly, we retrieve all the edges in $\mathcal{G}$ that connect any two nodes in $V$ to obtain $E \subseteq \mathcal{E}$, forming the final local KG, $G=(V,E)$.

\subsection{Graph initialization}
\label{app:setup:graph}

For the ConceptNet knowledge graph used in the general commonsense domain (\S \ref{sec:experiments}), we follow the method of MHGRN \cite{feng2020scalable} to prepare the initial KG node embeddings. Specifically, we convert triplets in the KG into sentences using pre-defined templates for each relation. Then, these sentences are fed into BERT-Large \cite{devlin2018bert} to compute embeddings for each sentence. Finally, for each entity, we collect all sentences containing the entity, extract all token representations of the entity’s mention spans in these sentences, and return the mean pooling of these representations.

For the UMLS knowledge graph used in the biomedical domain (\S \ref{sec:experiments-biomed}), node embeddings are initialized similarly using the pooled token output embeddings of the entity name from BioLinkBERT \cite{yasunaga2022linkbert}.

While extremely rare (<\,1\%), in case when the input text does not yield any linked entity, we represent the graph using a dummy node initialized with 0, i.e., \methodname backs off to only using the text side representations because the graph propagates no information.

\subsection{Hyperparameters}
\label{app:setup:hyperparameters}

\begin{table}[h]
\centering
\resizebox{\textwidth}{!}{%
\begin{tabular}{llcccc}
\toprule
\multirow{2}{*}{\textbf{Category}} & \multirow{2}{*}{\textbf{Hyperparameter}} & \multicolumn{2}{c}{\textbf{Commonsense domain}} & \multicolumn{2}{c}{\textbf{Biomedical domain}} \\
\cmidrule(lr){3-4} \cmidrule(lr){5-6} 
 &  & ~~\textbf{Pretrain}~~ & ~~\textbf{Finetune}~~ & ~~\textbf{Pretrain}~~ & ~~\textbf{Finetune}~~ \\ \midrule
\multirow{7}{*}{Model architecture} 
 & Number of text-KG fusion layers $M$ & 5 & 5 & 5 & 5 \\ \cmidrule(l){2-6} 
 & Number of Unimodal LM layers $N$ & 19 & 19 & 19 & 19 \\ \cmidrule(l){2-6} 
 & Number of attention heads in GNN & 2 & 2 & 2 & 2\\ \cmidrule(l){2-6} 
 & Dimension of node embeddings and the messages in GNN & 200 & 200 & 200 & 200 \\ \cmidrule(l){2-6} 
 & Dimension of MLP hidden layers (except MInt operator) & 200 & 200 & 200 & 200 \\ \cmidrule(l){2-6} 
 & Number of hidden layers of MLPs & 1 & 1 & 1 & 1 \\ \cmidrule(l){2-6}  
 & Dimension of MInt operator hidden layer & 400 & 400 & 400 & 400 \\ 
 \midrule
Regularization & Dropout rate of the embedding layer, GNN layers and dense layers & 0.2 & 0.2 & 0.2 & 0.2 \\ 
\midrule
\multirow{10}{*}{Optimization} & Learning rate of parameters in LM & 2e-5 & $\{$1e-5, 2e-5, 3e-5$\}$  & 2e-5 & $\{$1e-5, 2e-5, 3e-5$\}$ \\ \cmidrule(l){2-6} 
 & Learning rate of parameters not in LM & 3e-4 & $\{$3e-4, 1e-3$\}$ & 3e-4 & $\{$1e-4, 3e-4$\}$ \\ \cmidrule(l){2-6} 
 & Number of epochs in which LM's parameters are kept frozen & 2 & 4 & 2 & 4 \\ \cmidrule(l){2-6} 
 & Optimizer & RAdam & RAdam & RAdam & RAdam \\ \cmidrule(l){2-6}
 & Learning rate schedule
    & \begin{tabular}{@{}c@{}}linear warmup \\[-0.5mm] and decay\end{tabular} 
    & \begin{tabular}{@{}c@{}}linear warmup \\[-0.5mm] and decay\end{tabular}
    & \begin{tabular}{@{}c@{}}linear warmup \\[-0.5mm] and decay\end{tabular}
    & \begin{tabular}{@{}c@{}}linear warmup \\[-0.5mm] and decay\end{tabular} \\ \cmidrule(l){2-6}
 & Warmup ratio & 0.1 & 0.1 & 0.1 & 0.1 \\ \cmidrule(l){2-6}
 & Batch size & 8,192 & 128 & 8,192 & 128 \\ \cmidrule(l){2-6}
 & Number of epochs & - & 10--70 & - & 10--70 \\ \cmidrule(l){2-6}
 & Number of steps & 20,000 & - & 20,000 & - \\ \cmidrule(l){2-6}
 & Max gradient norm (gradient clipping) & 1.0 & 1.0 & 1.0 & 1.0 \\ \midrule
\multirow{2}{*}{Data} & Max number of nodes & 200 & 200 & 200 & 200 \\ \cmidrule(l){2-6}
 & Max number of tokens & 512 & $\{$128, 256$\}$ & 512 & 512 \\ \bottomrule
\end{tabular}%
}\vspace{2mm}
\caption{Hyperparameter settings for models and experiments}
\label{tab:hyperparameters}
\end{table}

\newpage

\subsection{Downstream evaluation tasks}
\label{app:setup:eval}

We use the following nine commonsense reasoning benchmarks for the experiments in the general domain (\S \ref{sec:experiments}).

\textbf{CommonsenseQA (CSQA)} \cite{talmor2018commonsenseqa} is a 5-way multiple-choice QA task testing commonsense reasoning. The dataset has 12,102 questions. We use the in-house data splits by \cite{lin2019kagnet}.

\textbf{OpenbookQA (OBQA)} \cite{obqa} is a 4-way multiple-choice QA task containing elementary science questions. It has 5,957 questions. We use the original data splits in \cite{mihaylov2018knowledgeable}.

\textbf{RiddleSense (Riddle)} \cite{lin2021riddlesense} is a 5-way multiple-choice task testing complex riddle-style commonsense reasoning. It has 5,715 questions. We split the dev set in half to make in-house dev/test sets.

\textbf{AI2 Reasoning Challenge, Challenge Set (ARC)} \cite{clark2018think} is a 4-way multiple-choice QA task containing science exam questions. It has 2,590 questions. We use the original data splits in \cite{clark2018think}.

\textbf{CosmosQA} \cite{huang2019cosmos} is a 4-way multiple-choice QA task testing commonsense reasoning with long narratives. It has 35.6K questions. We split the dev set in half to make in-house dev/test sets.

\textbf{HellaSwag} \cite{zellers2019hellaswag} is a 4-way multiple-choice task testing grounded commonsense reasoning about events. It has 70K questions. We split the dev set in half to make in-house dev/test sets.

\textbf{Physical Interaction QA (PIQA)} \cite{bisk2020piqa} is a 3-way multiple-choice QA task testing physics reasoning about objects. It has 20K questions. We split the dev set in half to make in-house dev/test sets.

\textbf{Social Interaction QA (SIQA)} \cite{sap2019socialiqa} is a 3-way multiple-choice QA task testing social commonsense reasoning. It has 37K questions. We use the original data splits in \cite{sap2019socialiqa}.

\textbf{Abductive Natural Language Inference (aNLI)} \cite{bhagavatula2019abductive} is a 2-way multiple-choice task testing abductive commonsense reasoning. It has 170K questions. We use the original data splits in \cite{bhagavatula2019abductive}.

For the experiments in the biomedical domain (\S \ref{sec:experiments-biomed}), we use the following three biomedical NLP and reasoning benchmarks.

\textbf{MedQA-USMLE (MedQA)} \cite{jin2021disease} is a 4-way multiple-choice task containing United States Medical License Exam questions. The dataset has 12,723 questions. We use the original data splits in \cite{jin2021disease}.

\textbf{PubMedQA} \cite{jin2019pubmedqa} is a 3-way multiple-choice task testing biomedical language understanding and reasoning. The dataset has 1,000 questions. We use the original data splits in \cite{jin2019pubmedqa}.

\textbf{BioASQ} \cite{nentidis2019results} is a 2-way multiple-choice task testing biomedical language understanding and reasoning. The dataset has 885 questions. We use the original data splits in \cite{nentidis2019results}.

\begin{table}[h]
    \centering
    \resizebox{\textwidth}{!}{
    \begin{tabular}{rl}
        \toprule
         \textbf{Dataset} & \textbf{Example} \\
         \midrule
         \multirow{2}{*}{CommonsenseQA} & A weasel has a thin body and short legs to easier burrow after prey in a what? \\
         & (A) tree (B) mulberry bush (C) chicken coop (D) viking ship {\color{blue} \textbf{(E) rabbit warren}} \\
         \midrule
         \multirow{3}{*}{OpenbookQA} & Which of these would let the 
most heat travel through?\\
        & (A) a new pair of jeans \:\:\:\:\:\:\:\:\:\:\,{\color{blue} \textbf{(B) a steel spoon in a cafeteria}}\\
& (C) a cotton candy at a store \:(D) a calvin klein cotton hat\\
        \midrule
        \multirow{2}{*}{RiddleSense} & What home entertainment equipment requires cable?\\
       & (A) radio shack (B) substation (C) cabinet \color{blue}{\textbf{(D) television}} (E) desk \\
         \midrule
         \multirow{2}{*}{AI2 Reasoning Challenge} & Which property of a mineral can be determined just by looking at it?\\
       & \color{blue}{\textbf{(A) luster}} (B) mass (C) weight (D) hardness \\
        \midrule
        \multirow{7}{*}{CosmosQA} &
          It's a very humbling experience when you need someone to dress you every morning, tie your shoes,\\
        & and put your hair up. Every menial task takes an unprecedented amount of effort. It made me\\
        & appreciate Dan even more. But anyway I shan't dwell on this  (I'm not dying after all) and not let\\
        & it detact from my lovely 5 days with my friends visiting from Jersey. What's a possible reason the\\
        & writer needed someone to dress him every morning?\\
        & (A) The writer doesn't like putting effort into these tasks. \color{blue}{\textbf{(B) The writer has a physical disability.}}\\
        & (C) The writer is bad at doing his own hair. ~~~~~~~~~~~~~~~~~~~~~~\,(D) None of the above choices.\\
        \midrule
        \multirow{5}{*}{HellaSwag} & 
          A woman is outside with a bucket and a dog. The dog is running around trying to avoid a bath. She\\
        & (A) rinses the bucket off with soap and blow dries the dog's head.\\
        & (B) uses a hose to keep it from getting soapy.\\
        & \color{blue}{\textbf{(C) gets the dog wet, then it runs away again.}}\\
        & (D) gets into the bath tub with the dog.\\
        \midrule
        \multirow{2}{*}{Physical Interaction QA} 
        & You need to break a window. Which object would you rather use?\\
        & \color{blue}{\textbf{(A) a metal stool}} (B) a giant bear (C) a bottle of water \\
       \midrule
        \multirow{3}{*}{Social Interaction QA} 
        & In the school play, Robin played a hero in the struggle to the death with the angry villain.\\ 
        & How would others feel as a result?\\
        & (A) sorry for the villain \color{blue}{\textbf{(B) hopeful that Robin will succeed}} (C) like Robin should lose the fight\\
        \midrule
        \multirow{4}{*}{aNLI} 
        & Obs1: It was a gorgeous day outside.\\
        & Obs2: She asked her neighbor for a jump-start.\\
        & \color{blue}{\textbf{Hyp1: Mary decided to drive to the beach, but her car would not start due to a dead battery.}}\\
        & Hyp2: It made a weird sound upon starting.\\
         \midrule
         \midrule
         \multirow{9}{*}{MedQA-USMLE} 
        & A 57-year-old man presents to his primary care physician with a 2-month history of right upper and\\ & lower extremity weakness. He noticed the weakness when he started falling far more frequently while\\
        & running errands. Since then, he has had increasing difficulty with walking and lifting objects.\\
        & His past medical history is significant only for well-controlled hypertension, but he says that some\\
        & members of his family have had musculoskeletal problems. His right upper extremity shows forearm\\
        & atrophy and depressed reflexes while his right lower extremity is hypertonic with a positive Babinski\\ 
        & sign. Which of the following is most likely associated with the cause of this patients symptoms? \\
        & (A) HLA-B8 haplotype \:\:\:\:\:\: (B) HLA-DR2 haplotype \\ 
        & {\color{blue} \textbf{(C) Mutation in SOD1}} \:\:\:\:\: (D) Mutation in SMN1 \\
        \midrule
        \multirow{9}{*}{PubMedQA} 
        & Recent studies have demonstrated that statins have pleiotropic effects, including anti-inflammatory\\ & effects and atrial fibrillation (AF) preventive effects [...]\\
        & 221 patients underwent CABG in our hospital from 2004 to 2007. 14 patients with preoperative AF and\\
        & 4 patients with concomitant valve surgery [...]\\
        & The overall incidence of postoperative AF was 26\%. Postoperative AF was significantly lower in the\\
        & Statin group compared with the Non-statin group (16\%versus 33\%, p=0.005). Multivariate analysis\\ 
        & demonstrated that independent predictors of AF [...]\\
        & Do preoperative statins reduce atrial fibrillation after coronary artery bypass grafting?\\
        & \color{blue}{\textbf{(A) yes}} (B) no (C) maybe\\
        \midrule
        \multirow{4}{*}{BioASQ} 
        & LT4 absorption is unchanged by concomitant metformin ingestion. It has been hypothesized that\\
        & metformin may suppress serum thyrotropin (TSH) concentrations by enhancing LT4 absorption or by\\
        & directly affecting the hypothalamic-pituitary axis. Does metformin interfere thyroxine absorption?\\
        & (A) yes \color{blue}{\textbf{(B) no}}\\
         \bottomrule
    \end{tabular}
    }
    \vspace{1mm}
    \caption{Example for each downstream task dataset used in this work.}
    \label{tab:datasets}
\end{table}

\section{Additional Experimental Results}
\label{app:results}

\begin{table}[h]
\centering
\small
\begin{tabular}{lc}
\toprule
\textbf{Method}         & \textbf{Hit@3}           \\
\midrule
DistMult (i.e., KG only) & 61.3\\
\methodname (i.e., KG + text) & \textbf{78.1}\\
\bottomrule
\end{tabular}\vspace{2mm}
\caption{\textbf{KG link prediction performance on ConceptNet}. In addition to the NLP tasks we mainly used for downstream evaluation, \methodname can also perform KG link prediction tasks in downstream. We find that \methodname (which uses retrieved text besides the KG) achieves improved performance on the KG link prediction task compared to the baseline DistMult model (which does not use text).
}
\vspace{-2mm}
\label{tab:kg_link_pred}
\end{table}

\end{document}